\documentclass{article}

\usepackage{microtype}
\usepackage{graphicx}
\usepackage{subcaption}
\usepackage{booktabs}
\usepackage{hyperref}

\usepackage[accepted]{icml2026}
\makeatletter
\renewcommand{\Notice@String}{%
  \textit{Presented at the $\mathit{43}^{rd}$ International Conference
  on Machine Learning Workshop on Resource-Adaptive Foundation Model
  Inference (AdaptFM)}, Seoul, South Korea, 2026.
  Copyright 2026 by the author(s).}
\makeatother
\usepackage{amsmath}
\usepackage{amssymb}
\usepackage{mathtools}
\usepackage{amsthm}
\usepackage{xcolor}
\usepackage{multirow}
\usepackage{float}
\usepackage{placeins}
\usepackage[capitalize,noabbrev]{cleveref}

\icmltitlerunning{AAAC: Activation-Aware Adaptive Codebooks for 4-bit LLM Quantization}

\begin{document}
\raggedbottom

\twocolumn[
  \icmltitle{AAAC: Activation-Aware Adaptive Codebooks for 4-bit LLM Weight Quantization}

  \icmlsetsymbol{equal}{*}

\begin{icmlauthorlist}
  \icmlauthor{Beshr IslamBouli}{equal,mit}
  \icmlauthor{David Jin}{equal,mit}
\end{icmlauthorlist}

  \icmlaffiliation{mit}{Massachusetts Institute of Technology, Cambridge, MA, USA}

\icmlcorrespondingauthor{Beshr IslamBouli}{beshr@mit.edu}
\icmlcorrespondingauthor{David Jin}{jindavid@mit.edu}

  \icmlkeywords{Quantization, LLM, Post-Training Quantization, Codebook Learning}

  \vskip 0.3in
]

\printAffiliationsAndNotice{\icmlEqualContribution}

\begin{abstract}
Post-training weight-only quantization to 4 bits is widely used to reduce the memory and compute costs of large language model inference. Existing PTQ methods, such as AWQ and GPTQ, improve how weights are mapped onto a fixed 4-bit grid through scaling, clipping, or error compensation. To further improve accuracy, methods such as OmniQuant and QuIP\# use gradient-assisted algorithms at the cost of hours of quantization time. In this work, we propose AAAC (Activation-Aware Adaptive Codebooks), a lightweight method for 4-bit LLM weight quantization. AAAC replaces the fixed scalar codebook used in standard quantization with two small learned scalar codebooks (64 bytes) per layer. Each group of weights selects the codebook that minimizes activation-weighted reconstruction error, encoding the choice in the unused sign bit of the group's positive scale and adding zero storage overhead. AAAC completes in 3--30 minutes on a single GPU, and adds no memory beyond the model itself. We evaluate against AWQ, GPTQ, IF4, GPTVQ, OmniQuant, SqueezeLLM, and QuIP\# across model families. AAAC outperforms baselines at orders-of-magnitude less quantization time.
\end{abstract}

\section{Introduction}
\label{sec:intro}
Large language models (LLMs) are widely deployed under memory and compute constraints that make weight quantization essential~\citep{touvron2023llama,qwen2025qwen25}. Weight-only 4-bit quantization has emerged as a practical inference setting: it reduces model storage while retaining acceptable quality for many models~\citep{dettmers2023spqr}.

Most post-training quantization (PTQ) methods improve how weights are mapped onto a fixed 4-bit grid. In group-wise quantization, each group of weights shares a scale $s$, and each normalized weight $\tilde{w}=w/s$ is rounded to a discrete code under a format-defined table $\mathcal{T}$. The same table is then used during dequantization:
\begin{equation}
\label{eq:pipeline}
c = Q(w;\,s), \qquad \hat{w} = Q^{-1}(c;\,s) = \mathcal{T}[c]\cdot s.
\end{equation}
For INT4, $\mathcal{T}$ is an integer grid; for NVFP4, $\mathcal{T}$ is the FP4 grid. Once the format is chosen, this scalar grid is usually treated as fixed. Methods such as AWQ and GPTQ therefore improve quantization around this fixed grid, using channel scaling~\citep{lin2024awq}, clipping, or error compensation~\citep{frantar2023gptq}, while leaving the scalar reconstruction grid itself unchanged. We argue that this fixed-grid assumption leaves quality on the table. The dequantized values, not the discrete codes alone, determine the weights actually used in W4A16 and W4A8 inference. A 4-bit format fixes the number of stored codes and the group-wise scale structure, but it need not fix the best scalar grid for a given layer. This suggests a complementary PTQ direction: instead of only asking how to map weights onto a fixed FP4 or INT4 table, learn adaptive scalar codebooks that better match the weight distribution and the activation statistics of the layer.

We propose \textbf{Activation-Aware Adaptive Codebooks (AAAC)}, a lightweight method for 4-bit LLM weight quantization. AAAC replaces the single fixed table $\mathcal{T}$ with two learned scalar codebooks per layer, $\mathcal{T}_0$ and $\mathcal{T}_1$. Each group of weights selects the codebook that minimizes activation-weighted reconstruction error, and weights in that group are stored as ordinary 4-bit indices into the selected codebook. AWQ and GPTQ improve how weights are mapped onto a fixed 4-bit table; AAAC improves the table used for that mapping.

AAAC distinguishes itself from prior PTQ methods on three axes.
\textbf{Negligible overhead}: AAAC adds only two BF16 codebooks (64 bytes) per layer. The per-group selection bit is stored in the unused sign of the always-positive scale, adding zero storage.
\textbf{No gradients}: codebooks are learned via activation-weighted $k$-means on forward-pass activations alone---no backpropagation, no Hessian or Fisher computation.
\textbf{No memory blowup}: AAAC processes each layer independently using only its activations and codebook statistics, so peak memory stays close to the model size itself. Across Llama and Qwen models in both NVFP4 and INT4 settings, we evaluate against AWQ, GPTQ, IF4, GPTVQ, OmniQuant, SqueezeLLM, and QuIP\#. AAAC outperforms baselines at orders-of-magnitude less quantization time.

\section{Related Work}
\label{sec:related}

Post-training quantization methods for LLMs can be divided into gradient-free approaches, which use only forward-pass calibration data, and gradient-assisted approaches, which incur larger quantization times and/or peak memory overhead.

\noindent\textbf{Gradient-free PTQ.}
GPTQ~\citep{frantar2023gptq} quantizes weights column-by-column using inverse-Hessian error compensation.
AWQ~\citep{lin2024awq} protects salient weights through activation-aware channel scaling.
IF4~\citep{if4} selects per-group between FP4 and INT4 codebooks based on MSE.
All three complete in minutes with peak memory close to the original model size.

\noindent\textbf{Gradient-assisted PTQ.}
OmniQuant~\citep{shao2023omniquant} learns per-channel clipping thresholds and equivalent transforms via block-wise backpropagation through the rounding operation, requiring 1--16 hours for 7B--70B models.

SqueezeLLM~\citep{kim2023squeezellm} learns per-channel non-uniform lookup tables via Fisher-weighted k-means, where the Fisher diagonal is computed by backpropagating through the full model.
While the algorithm runs in 11--80 minutes, the Fisher step drives peak memory to 2--4$\times$ model size (33--292\,GB for 7B--65B).

QuIP\#~\citep{tseng2024quipsharp} applies a Randomized Hadamard Transform for incoherence processing, then rounds to a fixed E8 lattice codebook via BlockLDLQ.
Without fine-tuning this is gradient-free and completes in under 10 GPU-hours for 70B.
An optional fine-tuning stage adds backpropagation through the full model, raising the cost to roughly 100 GPU-hours.

GPTVQ~\citep{vanbaalen2024gptvq} extends GPTQ to vector quantization, learning per-group codebooks via Hessian-weighted EM followed by gradient descent on the layer-wise reconstruction objective to refine codebook entries.
Quantization takes 0.5--11 hours for 7B--70B models.

Our experiments show that gradient-assisted methods consistently outperform gradient-free methods, suggesting that gradients---with their associated quantization time and memory overhead---are necessary to close the gap to full-precision quality.
AAAC challenges this conclusion: it is a lightweight, gradient-free method that learns only two codebooks per layer (64 bytes overhead) via activation-weighted k-means.
Despite this simplicity, AAAC outperforms all gradient-free methods and matches or exceeds gradient-assisted methods, with 3--30 minutes of quantization time and peak memory no larger than the original model size.

\section{Method}
\label{sec:method}

\subsection{Background: Group-wise Quantization}
\label{sec:prelim}

Consider a linear layer computing $\mathbf{y}=\mathbf{x}\mathbf{W}^{\top}$ with $\mathbf{W}\in\mathbb{R}^{N\times K}$. In group-wise quantization, each row of $\mathbf{W}$ is partitioned into groups of $g$ contiguous elements. Each group shares a scale $s>0$, and each weight is represented by a discrete code.

\paragraph{Quantization.}
A scale $s$ is computed from the group's weight statistics. Each weight $w$ is normalized by $s$, rounded to the nearest entry in the format-defined table $\mathcal{T}$, and stored as a $b$-bit code:
\begin{equation}
\label{eq:base_quant}
\tilde{w}=w/s, \qquad c=\arg\min_i |\tilde{w}-\mathcal{T}[i]|.
\end{equation}

\paragraph{Dequantization.}
During inference, the stored code is looked up in the same table and multiplied by the scale:
\begin{equation}
\label{eq:base_dequant}
\hat{w}=\mathcal{T}[c]\cdot s.
\end{equation}
Thus the table $\mathcal{T}$ plays two roles: it defines the nearest-neighbor code assignment during quantization, and it defines the reconstruction value used during dequantization.

It is useful to distinguish the stored code from the reconstructed value.
For any scalar table $\mathcal{A}$, define the nearest-entry reconstruction operator
\begin{equation}
\label{eq:nearest_recon}
\operatorname{recon}_{\mathcal{A}}(\tilde{w})=\mathcal{A}\!\left[\arg\min_i |\tilde{w}-\mathcal{A}[i]|\right].
\end{equation}
This operator first finds the nearest entry of $\mathcal{A}$ to the normalized weight $\tilde{w}$, then returns the corresponding table value. Under the base format, the reconstructed normalized weight is $\operatorname{recon}_{\mathcal{T}}(\tilde{w})$, and the dequantized weight after quantization is $\operatorname{recon}_{\mathcal{T}}(\tilde{w})\cdot s$.

The table $\mathcal{T}$ is fixed by the format. For NVFP4, $\mathcal{T}$ contains the signed E2M1 values used by the FP4 format, such as $\{-6,-4,-3,\ldots,0,\ldots,3,4,6\}$. For INT4, $\mathcal{T}$ denotes the signed or zero-point-adjusted integer grid used by the implementation, expressed in normalized units. Thus both NVFP4 and INT4 can be written as group-wise scalar quantizers with a finite code alphabet and a per-group scale.

\paragraph{Key observation.}
Existing PTQ methods such as AWQ and GPTQ primarily improve quantization around the fixed table $\mathcal{T}$. They change the weight distribution, scales, or compensation procedure so that the fixed table produces better reconstructed values. AAAC targets a different degree of freedom: it learns the scalar table itself.

\subsection{AAAC: Activation-Aware Adaptive Codebooks}
\label{sec:aaac}

AAAC replaces the single fixed table $\mathcal{T}$ with two learned scalar codebooks $\mathcal{T}_0$ and $\mathcal{T}_1$. Each learned table has the same number of entries as the base 4-bit format, but its entries are optimized from calibration data rather than fixed by the format. A group of $S$ consecutive weights selects one of the two tables. The selected table defines the nearest-entry reconstruction used for that group.

\paragraph{Granularity parameters.}
AAAC has two granularity parameters. The \emph{scale group size} $g$ is fixed by the quantization format and determines how many weights share one scale: $g{=}16$ for NVFP4 and $g{=}128$ for INT4. The \emph{selection group size} $S$ determines how many weights share one codebook choice. Thus, every $S$ weights store one selection bit $\sigma$, and every $g$ weights store one scale.

When $S{=}g$, the selection group coincides with the scale group: each scale needs exactly one codebook-selection bit. Since the scale is always positive, its sign bit is unused, and we store $\sigma$ there with no additional selection storage. When $S{<}g$, the selection metadata costs $1/S$ bits per weight. Smaller $S$ gives finer codebook adaptation. We use $S{=}g{=}16$ for NVFP4 and evaluate $S\in\{16,128\}$ for INT4.

\paragraph{Modified quantization and dequantization.}
For a weight $w$ in selection group $j$, let $\tilde{w}=w/s$. AAAC selects table $\mathcal{T}_{\sigma_j}$ and reconstructs the normalized weight using the nearest-entry reconstruction operator:
\begin{equation}
\label{eq:aaac_dequant}
\hat{w}=\operatorname{recon}_{\mathcal{T}_{\sigma_j}}(\tilde{w})\cdot s.
\end{equation}
Equivalently, the stored 4-bit code is the nearest-entry index under the selected table:
\begin{equation}
\label{eq:aaac_code}
c=\arg\min_i|\tilde{w}-\mathcal{T}_{\sigma_j}[i]|.
\end{equation}
Thus AAAC preserves the scalar 4-bit code width, group-wise scale structure, and packed representation, but it does not require the code assignment to match round-to-nearest under the original table $\mathcal{T}$. Its improvement comes from learning better scalar reconstruction grids and choosing, per group, which grid to use.

\paragraph{Activation importance.}
For a linear layer $\mathbf{y}=\mathbf{x}\mathbf{W}^{\top}$, a perturbation $\delta\mathbf{W}$ produces output error $\delta\mathbf{y}=\mathbf{x}\delta\mathbf{W}^{\top}$. The expected squared output error contributed by column $k$ of $\delta\mathbf{W}$ is proportional to $\mathbb{E}[x_k^2]\|\delta\mathbf{W}_{:,k}\|^2$. Summing over a calibration set $\mathbf{X}\in\mathbb{R}^{T\times K}$ gives the per-column importance
\begin{equation}
\label{eq:importance}
I_k=\sum_{t=1}^{T}X_{t,k}^2.
\end{equation}
Minimizing $\sum_k I_k\|\delta\mathbf{W}_{:,k}\|^2$ is therefore a diagonal proxy for minimizing calibration output error. Columns with larger activation energy receive larger weight in the reconstruction objective.

\paragraph{Table selection.}
Consider a selection group $j$ containing normalized weights $\{\tilde{w}_{j,1},\ldots,\tilde{w}_{j,S}\}$ and activation importances $\{I_{j,1},\ldots,I_{j,S}\}$. AAAC assigns the group to the table that minimizes activation-weighted reconstruction error:
\begin{equation}
\label{eq:selection}
\sigma_j=\arg\min_{r\in\{0,1\}}\sum_{k=1}^{S}I_{j,k}\left(\tilde{w}_{j,k}-\operatorname{recon}_{\mathcal{T}_r}(\tilde{w}_{j,k})\right)^2.
\end{equation}
The table choice and code assignment are coupled: each candidate table defines both the reconstruction values and the nearest-entry regions used to encode the weights in the group.

\paragraph{Storage.}
Each learned table contains $|\mathcal{T}|$ values stored in BF16, for a total of $2|\mathcal{T}|\cdot2$ bytes per layer, which is at most 64 bytes. For a 7B model with ${\sim}224$ linear layers, this is about 14\,KB, negligible compared with the model weights. The selection bits $\{\sigma_j\}$ require no additional storage when $S{=}g$ if they are stored in the unused sign bit of each positive scale, and require $1/S$ bits per weight when $S{<}g$.

\subsection{Learning Algorithm}
\label{sec:algorithm}

We now describe how the tables $\mathcal{T}_0$ and $\mathcal{T}_1$ are obtained from calibration data. The two tables are learned per layer using a small calibration set. Learning alternates between assigning selection groups to tables and refining the table entries. Algorithm~\ref{alg:aaac} provides pseudocode.

\begin{algorithm}[!t]
\caption{AAAC: Learning activation-aware adaptive codebooks for one layer}
\label{alg:aaac}
\begin{algorithmic}[1]
\REQUIRE Weight matrix $\mathbf{W}\in\mathbb{R}^{N\times K}$, calibration activations $\mathbf{X}\in\mathbb{R}^{T\times K}$
\REQUIRE Group sizes $g$ (scale), $S$ (selection); iterations $n_{\mathrm{outer}}$, $n_{\mathrm{inner}}$
\STATE Compute scales $s$ per group of $g$ weights
\STATE Normalize weights: $\tilde{\mathbf{W}}\leftarrow \mathbf{W}/s$
\STATE Compute activation importance: $I_k\leftarrow \sum_t X_{t,k}^2$ for each column $k$
\STATE Initialize $\mathcal{T}_0$ from full-range quantiles of $\tilde{\mathbf{W}}$ \COMMENT{Eq.~\ref{eq:init_t0}}
\STATE Initialize $\mathcal{T}_1$ from half-shifted quantiles of $\tilde{\mathbf{W}}$ \COMMENT{Eq.~\ref{eq:init_t1}}
\FOR{$t=1$ to $n_{\mathrm{outer}}$}
    \STATE \textbf{// Assignment step}
    \FOR{each selection group $j$}
        \STATE $\sigma_j \leftarrow \arg\min_{r\in\{0,1\}}$
        \STATE \hspace{2em} $\sum_{k=1}^{S} I_{j,k}\left(\tilde{w}_{j,k}-\operatorname{recon}_{\mathcal{T}_r}(\tilde{w}_{j,k})\right)^2$
    \ENDFOR
    \STATE \textbf{// Table update step}
    \FOR{$r\in\{0,1\}$}
        \STATE Let $\mathcal{W}_r$ be the multiset of pairs $(\tilde{w}_{j,k},I_{j,k})$ with $\sigma_j=r$
        \FOR{$\ell=1$ to $n_{\mathrm{inner}}$}
            \FOR{each entry $i$ in $\mathcal{T}_r$}
                \STATE Let $\mathcal{V}_{r,i}\subseteq\mathcal{W}_r$ be the Voronoi cell of entry $i$
                \STATE $\mathcal{T}_r[i]\leftarrow \dfrac{\sum_{(\tilde{w},I)\in\mathcal{V}_{r,i}} I\tilde{w}}{\sum_{(\tilde{w},I)\in\mathcal{V}_{r,i}} I}$
                \STATE If the denominator is zero, keep the previous value of $\mathcal{T}_r[i]$
            \ENDFOR
        \ENDFOR
        \STATE Sort $\mathcal{T}_r$ in ascending order
    \ENDFOR
\ENDFOR
\STATE Round $\mathcal{T}_0,\mathcal{T}_1$ to BF16 precision
\STATE Recompute $\{\sigma_j\}$ using Eq.~\ref{eq:selection}
\STATE Pack 4-bit codes as nearest-entry indices into $\mathcal{T}_{\sigma_j}$
\STATE Store $\{\sigma_j\}$ in sign bits of $s$ if $S{=}g$, or separately if $S{<}g$
return $\mathcal{T}_0,\mathcal{T}_1$, packed 4-bit codes, scales, and $\{\sigma_j\}$
\end{algorithmic}
\end{algorithm}

\paragraph{Initialization.}
Let $M=|\mathcal{T}|$ be the number of entries in each table, and let $\mathrm{Quantile}_{q}(\{\tilde{w}\})$ denote the $q$-quantile of the normalized weight distribution. We initialize $\mathcal{T}_0$ using $M$ evenly spaced quantiles spanning the full range:
\begin{equation}
\label{eq:init_t0}
\mathcal{T}_0[i]=\mathrm{Quantile}_{i/(M-1)}(\{\tilde{w}\}), \qquad i=0,\ldots,M-1.
\end{equation}
We initialize $\mathcal{T}_1$ using $M$ evenly spaced quantiles starting at a half-step offset and ending at the maximum:
\begin{equation}
\label{eq:init_t1}
\mathcal{T}_1[i]=\mathrm{Quantile}_{\delta+(1-\delta)i/(M-1)}(\{\tilde{w}\}), \qquad \delta=\frac{1}{2(M-1)}.
\end{equation}
This shifts the second grid forward by half a quantile spacing, giving the two tables distinct but nearby starting points. Results are insensitive to the exact offset.

\paragraph{Alternating optimization.}
Learning alternates between two steps for $n_{\mathrm{outer}}$ iterations. In the \emph{assignment step}, each selection group is assigned to the table that minimizes its activation-weighted reconstruction error, as in Eq.~\ref{eq:selection}. In the \emph{table update step}, each table is refined by $n_{\mathrm{inner}}$ iterations of activation-weighted scalar $k$-means. For each table, normalized weights assigned to that table are assigned to their nearest table entry, and each entry is updated to the activation-weighted centroid of its Voronoi cell:
\begin{equation}
\label{eq:kmeans_update}
\mathcal{T}_{r}[i]\leftarrow \frac{\sum_{(\tilde{w},I)\in\mathcal{V}_{r,i}} I\tilde{w}}{\sum_{(\tilde{w},I)\in\mathcal{V}_{r,i}} I}.
\end{equation}
Here, $\mathcal{V}_{r,i}$ is the set of normalized weights assigned to entry $i$ of table $\mathcal{T}_r$. If a Voronoi cell is empty, we keep the entry at its previous value. After the final inner update, the entries of each table are sorted in ascending order. The outer loop matters more than the inner loop: reassigning groups to tables has a larger effect on quality than refining table entries within a fixed assignment. We find empirically that a single outer iteration already achieves 82\% of the total improvement, so the overall cost of the procedure is dominated by the assignment step rather than the inner $k$-means refinement.

\paragraph{Computational cost.}
Each outer iteration requires two forward-only passes over the layer's weights—one for group assignment, one for $k$-means updates—with no gradients, no Hessian-vector products, and no backpropagation. The algorithm processes layers independently, so peak GPU memory is dominated by the model itself plus a per-layer activation buffer. This enables in-memory calibration of a 27B model in roughly 10 minutes ($n_{\mathrm{outer}}{=}3$, $n_{\mathrm{inner}}{=}10$).

\section{Experiments}
\label{sec:experiments}

\subsection{Setup}
\label{sec:setup}

\paragraph{Models.}
We evaluate on models spanning Llama (3.2-1B, 3.2-3B, 3.1-8B)~\citep{grattafiori2024llama3}, Qwen2.5 (0.5B, 7B, 14B, 32B)~\citep{qwen2025qwen25}, and Qwen3.5 (2B, 4B, 9B, 27B) ~\citep{qwen2026qwen35}. To compare with numbers reported by prior methods, we evaluate on LLaMA-1 (7B, 13B) and LLaMA-2 (7B, 13B)~\citep{touvron2023llama}.

\paragraph{Evaluation.}
We report perplexity on WikiText-2~\citep{merity2016wikitext} and C4~\citep{raffel2020c4}. For WikiText-2, we follow the GPTQ protocol: the full test set is tokenized without filtering and split into non-overlapping 2048-token segments, with per-token cross-entropy computed via manual logit shifting~\citep{frantar2023gptq}. For C4, we use the validation split with up to 256 segments. We additionally evaluate on downstream tasks using lm-evaluation-harness~\citep{eval-harness}.

\paragraph{Gap recovery.} To summarize the overall effectiveness of each method, we report \emph{gap recovery}: the percentage of the quantization-induced degradation that a method eliminates relative to the round-to-nearest baseline. For perplexity (lower is better), gap recovery is defined as $(\text{PPL}_{\text{RTN}} - \text{PPL}_{\text{method}}) / (\text{PPL}_{\text{RTN}} - \text{PPL}_{\text{full}}) \times 100\%$, where $\text{PPL}_{\text{RTN}}$ is the round-to-nearest perplexity and $\text{PPL}_{\text{full}}$ is the full-precision baseline. For accuracy (higher is better), the analogous formula is $(\text{Acc}_{\text{method}} - \text{Acc}_{\text{RTN}}) / (\text{Acc}_{\text{full}} - \text{Acc}_{\text{RTN}}) \times 100\%$. A recovery of 0\% means no improvement over RTN; 100\% means full-precision quality is restored.

\paragraph{AAAC configurations.}
A smaller selection granularity $S$ allows AAAC to better adapt to local weight structure within each layer. The NVFP4 memory layout groups weights into blocks of 16 with shared FP8 scales, and this $g{=}16$ setting is ideal for AAAC: 16 weights is small enough for fine-grained codebook selection, and the selection bit can be stored in the scale sign at zero overhead. IF4 uses the same $g{=}16$ layout, and AWQ can be straightforwardly adapted to it. We report $g{=}16$ results alongside both methods. To compare with prior methods that use the standard INT4 layout with $g{=}128$, we also report $S{=}g{=}128$ (zero overhead). However, $S{=}128$ is too coarse for effective codebook selection, so we additionally report $S{=}16$, adding a negligible $1/16{=}0.0625$\,bpw overhead. We use $n_{\mathrm{outer}}{=}3$, $n_{\mathrm{inner}}{=}10$ with 4 calibration sequences from C4. We used a H100 GPU.

\begin{table}[t]
\caption{$g{=}16$ W4A16 perplexity on WikiText-2 ($\downarrow$). AAAC uses $S{=}g{=}16$ (zero overhead). Best quantized result in \textbf{bold}.}
\label{tab:g16_wiki}
\centering
\small
\begin{tabular}{l c c c c c}
\toprule
Model & BF16 & RTN & IF4 & AWQ & AAAC \\
\midrule
Llama-3.2-1B   & 9.71 & 10.68 & 10.61 & 10.46 & \textbf{10.21} \\
Llama-3.2-3B   & 7.77 &  8.19 &  8.13 &  8.11 & \textbf{7.99} \\
Llama-3.1-8B   & 6.19 &  6.56 &  6.51 &  6.51 & \textbf{6.39} \\
\midrule
Qwen3.5-2B     & 12.06 & 13.21 & 12.70 & 12.56 & \textbf{12.37} \\
Qwen3.5-4B     & 9.42 & 10.25 & 10.21 & 10.16 & \textbf{9.92} \\
Qwen3.5-9B     & 8.51 &  9.11 &  9.38 &  8.91 & \textbf{8.56} \\
Qwen3.5-27B    & 6.80 &  7.02 &  6.92 &  7.03 & \textbf{6.88} \\
\midrule
\emph{Average}  & \emph{8.64} & \emph{9.29} & \emph{9.21} & \emph{9.11} & \emph{\textbf{8.90}} \\
\emph{Recovery}  & \emph{---} & \emph{---} & \emph{12.3\%} & \emph{28.1\%} & \emph{\textbf{59.2\%}} \\
\bottomrule
\end{tabular}
\end{table}

\FloatBarrier
\paragraph{Baselines and reproducibility.}
We compare against round-to-nearest (RTN), IF4~\citep{if4}, AWQ~\citep{lin2024awq}, GPTQ~\citep{frantar2023gptq}, GPTVQ~\citep{vanbaalen2024gptvq}, OmniQuant~\citep{shao2023omniquant}, SqueezeLLM~\citep{kim2023squeezellm}, and QuIP\#~\citep{tseng2024quipsharp}. GPTQ numbers are taken from~\citet{lin2024awq}. We implement IF4 following~\citet{if4}: for each group, both FP4 and scaled INT4 quantizations are evaluated, and the option with lower MSE is selected. For AWQ at $g{=}128$, we use the official AutoAWQ package; for $g{=}16$, we implement their per-channel scaling algorithm ourselves as AutoAWQ does not support FP4-based quantization. GPTVQ, OmniQuant, SqueezeLLM, and QuIP\# numbers are taken from their respective papers; we verify that our RTN and full-precision baselines match those reported, ensuring consistent evaluation protocols.

\subsection{Comparison with Gradient-Free Methods}
\label{sec:gradient_free}

\subsubsection{Group Size 16 (NVFP4 Memory Layout)}
\label{sec:g16_results}

The NVFP4 memory layout stores weights in groups of 16 with shared FP8 scales, and several recent methods build on this layout. IF4~\citep{if4} selects per-group between the native FP4 and a scaled INT4 codebook. AWQ~\citep{lin2024awq}, originally designed for INT4, can be straightforwardly adapted to this layout. AAAC uses $S{=}g{=}16$, storing the selection bit in the scale sign at zero overhead. Table~\ref{tab:g16_wiki} presents WikiText-2 perplexity across 7 Llama and Qwen3.5 models.

\subsubsection{Group Size 128 (INT4 Layout)}
\label{sec:g128_results}

To compare with prior methods that use the standard INT4 layout with $g{=}128$, we evaluate AAAC at $S{=}g{=}128$ (zero overhead) and $S{=}16$ (+0.0625\,bpw).
Table~\ref{tab:int4_awq} compares AAAC against AWQ on Llama-3 and Qwen2.5 models, using the official AutoAWQ package.
With $S{=}g{=}128$ (zero overhead), AAAC already outperforms AWQ on the majority of models. The remaining cases where AWQ leads are attributable to the coarse selection granularity: with only one codebook choice per 128 weights, the method cannot adapt to local variations within each group. With $S{=}16$, adding only 0.0625\,bpw, AAAC consistently outperforms AWQ on all models.
Table~\ref{tab:int4_gptq} compares AAAC against GPTQ on LLaMA-1 and LLaMA-2 models. GPTQ numbers are taken from~\citet{lin2024awq}. AAAC at $S{=}16$ outperforms GPTQ on all four models. Even at $S{=}128$ (zero overhead), AAAC matches or outperforms GPTQ on all four models.

\begin{table}[t]
\caption{$g{=}128$ W4A16 WikiText-2 perplexity ($\downarrow$) vs.\ AWQ. AWQ uses the official AutoAWQ package. \textbf{Bold}: best result per model.}
\label{tab:int4_awq}
\centering
\small
\begin{tabular}{l c c c c c}
\toprule
 & & & & \multicolumn{2}{c}{AAAC} \\
\cmidrule(l){5-6}
Model & FP16 & RTN & AWQ & $S{=}128$ & $S{=}16$ \\
\midrule
Llama-3.2-3B  &  7.81 &  8.49 &  8.26 &  8.16 & \textbf{8.10} \\
Llama-3.1-8B  &  6.24 &  6.82 &  6.65 &  6.68 & \textbf{6.56} \\
Qwen2.5-0.5B  & 13.07 & 15.54 & 15.01 & 14.47 & \textbf{14.27} \\
Qwen2.5-7B    &  6.85 &  7.23 &  7.09 &  7.13 & \textbf{7.07} \\
Qwen2.5-14B   &  5.29 &  5.78 &  5.69 &  5.65 & \textbf{5.60} \\
\midrule
\emph{Average} & \emph{7.85} & \emph{8.77} & \emph{8.54} & \emph{8.42} & \emph{\textbf{8.32}} \\
\emph{Recovery} & \emph{---} & \emph{---} & \emph{25.0\%} & \emph{38.0\%} & \emph{\textbf{48.9\%}} \\
\bottomrule
\end{tabular}
\end{table}

\begin{table}[t]
\caption{$g{=}128$ W4A16 WikiText-2 perplexity ($\downarrow$) vs.\ GPTQ. GPTQ numbers from~\citet{lin2024awq}. \textbf{Bold}: best result per model.}
\label{tab:int4_gptq}
\centering
\small
\begin{tabular}{l c c c c c}
\toprule
 & & & & \multicolumn{2}{c}{AAAC} \\
\cmidrule(l){5-6}
Model & FP16 & RTN & GPTQ & $S{=}128$ & $S{=}16$ \\
\midrule
LLaMA-1-7B  & 5.68 & 5.96 & 5.85 & 5.82 & \textbf{5.76} \\
LLaMA-1-13B & 5.09 & 5.25 & 5.23 & 5.21 & \textbf{5.17} \\
LLaMA-2-7B  & 5.47 & 5.72 & 5.69 & 5.62 & \textbf{5.58} \\
LLaMA-2-13B & 4.88 & 4.98 & 4.98 & 4.98 & \textbf{4.94} \\
\midrule
\emph{Average} & \emph{5.28} & \emph{5.48} & \emph{5.44} & \emph{5.41} & \emph{\textbf{5.36}} \\
\emph{Recovery} & \emph{---} & \emph{---} & \emph{20.0\%} & \emph{35.0\%} & \emph{\textbf{60.0\%}} \\
\bottomrule
\end{tabular}
\end{table}

\subsection{Comparison with Gradient-Assisted Methods}
\label{sec:gradient_assisted}

\begin{table*}[t]
\caption{$g{=}128$ W4A16 WikiText-2 perplexity ($\downarrow$) vs.\ gradient-assisted methods. AAAC uses $S{=}16$. Best quantized result per row in \textbf{bold}.}
\label{tab:gradient_assisted}
\centering
\small
\setlength{\tabcolsep}{3.5pt}
\begin{tabular}{l c c c c c c c c}
\toprule
Model & FP16 & RTN & OmniQ & SqLLM & GPTVQ & QuIP\# & QuIP\#-FT & AAAC \\
\midrule
LLaMA-1-7B  & 5.68 & 5.96 & 5.77 & 5.77 & 5.96 & 5.83 & \textbf{5.76} & \textbf{5.76} \\
LLaMA-1-13B & 5.09 & 5.25 & \textbf{5.17} & \textbf{5.17} & \textbf{5.15} & 5.20 & \textbf{5.17} & \textbf{5.17} \\
LLaMA-2-7B  & 5.47 & 5.72 & 5.58 & 5.57 & 5.62 & 5.66 & \textbf{5.56} & 5.58 \\
LLaMA-2-13B & 4.88 & 4.98 & 4.95 & 4.96 & 4.97 & 5.00 & 4.95 & \textbf{4.94} \\
\midrule
\emph{Average} & \emph{5.28} & \emph{5.48} & \emph{5.37} & \emph{5.37} & \emph{5.43} & \emph{5.42} & \emph{5.36} & \emph{\textbf{5.36}} \\
\emph{Recovery} & \emph{---} & \emph{---} & \emph{55.0\%} & \emph{55.0\%} & \emph{25.0\%} & \emph{30.0\%} & \emph{60.0\%} & \emph{\textbf{60.0\%}} \\
\bottomrule
\end{tabular}
\end{table*}

\begin{table*}[t]
\caption{Combining AWQ and AAAC on WikiText-2 ($\downarrow$). AAAC uses $S{=}g$ (zero overhead). Best quantized result per row in \textbf{bold}.}
\label{tab:stacking}
\centering
\small
\begin{tabular}{l l c c c c c}
\toprule
$g$ & Model & Full & RTN & AWQ & AAAC & AWQ+AAAC \\
\midrule
128 & LLaMA-1-7B   & 5.68 & 5.96 & 5.81 & 5.82 & \textbf{5.80} \\
128 & LLaMA-1-13B  & 5.09 & 5.25 & 5.20 & 5.21 & \textbf{5.18} \\
128 & LLaMA-2-7B   & 5.47 & 5.72 & \textbf{5.62} & \textbf{5.62} & \textbf{5.62} \\
128 & LLaMA-2-13B  & 4.88 & 4.98 & \textbf{4.97} & 4.98 & 4.98 \\
16  & Qwen3.5-4B   & 9.42 & 10.25 & 10.16 & 9.92 & \textbf{9.52} \\
16  & Qwen3.5-9B   & 8.51 &  9.11 &  8.91 & \textbf{8.56} &  8.67 \\
16  & Qwen3.5-27B  & 6.80 &  7.02 &  7.03 &  6.88 & \textbf{6.80} \\
\midrule
\multicolumn{2}{l}{\emph{Average}} & \emph{6.55} & \emph{6.90} & \emph{6.81} & \emph{6.71} & \emph{\textbf{6.65}} \\
\multicolumn{2}{l}{\emph{Recovery}} & \emph{---} & \emph{---} & \emph{24.2\%} & \emph{53.3\%} & \emph{\textbf{70.5\%}} \\
\bottomrule
\end{tabular}
\end{table*}

As discussed in Section~\ref{sec:related}, gradient-assisted PTQ methods achieve strong results but at significant cost: OmniQuant requires 1--16 hours of block-wise backpropagation, SqueezeLLM requires 2--4$\times$ model size in peak memory for Fisher computation, GPTVQ requires 0.5--11 hours with gradient-based codebook refinement, and QuIP\#-FT requires up to 100 GPU-hours of end-to-end fine-tuning. Even QuIP\# without fine-tuning takes up to 10 GPU-hours. Table~\ref{tab:gradient_assisted} compares AAAC against all four methods.

\subsection{Combining AAAC with AWQ}
\label{sec:stacking}

AWQ modifies the weights before quantization via channel scaling, while AAAC learns the codebook used during quantization. Since they operate on different components, the two methods can be combined: AWQ first scales the weights, then AAAC learns codebooks on the scaled distribution. Table~\ref{tab:stacking} shows that AWQ+AAAC recovers 70.5\% of the quantization gap on average, compared to 53.3\% for AAAC alone.

\FloatBarrier
\subsection{Downstream Task Evaluation}
\label{sec:downstream}

To complement the perplexity results, we evaluate on 8 downstream tasks (BBH, MuSR, MMLU, GPQA, HellaSwag, ARC-C, PIQA, MATH) using Qwen3.5-9B and Qwen3.5-27B at $g{=}16$. AAAC achieves the highest average accuracy among all quantized methods (64.21\%) compared with the BF16 baseline (64.59\%). Full per-task results are in Appendix~\ref{app:downstream}.

\subsection{Additional Analysis}
\label{sec:additional}

\paragraph{W4A8 inference.}
Although AAAC uses activations to weight importance during calibration, it is not sensitive to activation precision at inference time. The same quantized model can serve both W4A16 and W4A8 inference without re-calibration. We verify this in Appendix~\ref{app:w4a8}, where the same $g{=}128$ models are evaluated with FP8 activations; AAAC's improvements carry over.

\paragraph{Sensitivity to hyperparameters.}
AAAC has few tunable factors: the selection group size $S$, the iteration budget, and the calibration set size. At $g{=}16$, $S{=}g{=}16$ is the natural choice at zero overhead. At $g{=}128$, $S{=}128$ (zero overhead) already outperforms AWQ on the majority of models (Tables~\ref{tab:int4_awq}--\ref{tab:int4_gptq}), while reducing $S$ to 16 (+0.0625\,bpw) consistently improves further. The algorithm is largely insensitive to the iteration budget: a fast configuration ($n_{\mathrm{outer}}{=}3$, $n_{\mathrm{inner}}{=}10$, under 2 minutes on an H100 for a 7B model) matches or trails something bigger like ($n_{\mathrm{outer}}{=}10$, $n_{\mathrm{inner}}{=}30$) by 0.01 perplexity. Calibration data requirements are similarly minimal: in our experience, 1 sequence (2{,}048 tokens) yields perplexity within 0.01 of the result with 4 or more sequences. We use 4 sequences as a default.



\section{Conclusion}
\label{sec:conclusion}

We have presented AAAC (Activation-Aware Adaptive Codebooks), a lightweight post-training quantization method that learns two small scalar codebooks per layer (64 bytes) via activation-weighted k-means on a small calibration set. By encoding the per-group codebook selection in the unused sign bit of the scale factor, AAAC adds zero storage cost when the selection and scale group sizes match. The method requires no gradients and adds no peak memory beyond the original model size. Despite this simplicity, AAAC consistently outperforms all gradient-free baselines and matches gradient-assisted methods that require hours of optimization and higher memory. We believe that learned codebooks at this minimal scale represent a practical direction for quantized language models.

\section*{Impact Statement}

This paper presents work whose goal is to advance the field of Machine Learning. There are many potential societal consequences of our work, none of which we feel must be specifically highlighted here.

\section*{Acknowledgements}
We thank Han Guo, Tarushii Goel, and Yoon Kim for helpful discussions.

\bibliography{refs}

@misc{dettmers2023spqr,
      title={SpQR: A Sparse-Quantized Representation for Near-Lossless LLM Weight Compression},
      author={Tim Dettmers and Ruslan Svirschevski and Vage Egiazarian and Denis Kuznedelev and Elias Frantar and Saleh Ashkboos and Alexander Borzunov and Torsten Hoefler and Dan Alistarh},
      year={2023},
      eprint={2306.03078},
      archivePrefix={arXiv},
      primaryClass={cs.CL},
      url={https://arxiv.org/abs/2306.03078},
}

@misc{frantar2023gptq,
      title={GPTQ: Accurate Post-Training Quantization for Generative Pre-trained Transformers},
      author={Elias Frantar and Saleh Ashkboos and Torsten Hoefler and Dan Alistarh},
      year={2023},
      eprint={2210.17323},
      archivePrefix={arXiv},
      primaryClass={cs.LG},
      url={https://arxiv.org/abs/2210.17323},
}

@misc{grattafiori2024llama3,
      title={The Llama 3 Herd of Models},
      author={Aaron Grattafiori and Abhimanyu Dubey and Abhinav Jauhri and Abhinav Pandey and Abhishek Kadian and Ahmad Al-Dahle and Aiesha Letman and Akhil Mathur and Alan Schelten and Alex Vaughan and Amy Yang and Angela Fan and Anirudh Goyal and Anthony Hartshorn and Aobo Yang and Archi Mitra and Archie Sravankumar and Artem Korenev and Arthur Hinsvark and Arun Rao and Aston Zhang and Aurelien Rodriguez and Austen Gregerson and Ava Spataru and Baptiste Roziere and Bethany Biron and Binh Tang and Bobbie Chern and Charlotte Caucheteux and Chaya Nayak and Chloe Bi and Chris Marra and Chris McConnell and Christian Keller and Christophe Touret and Chunyang Wu and Corinne Wong and Cristian Canton Ferrer and Cyrus Nikolaidis and Damien Allonsius and Daniel Song and Danielle Pintz and Danny Livshits and Danny Wyatt and David Esiobu and Dhruv Choudhary and Dhruv Mahajan and Diego Garcia-Olano and Diego Perino and Dieuwke Hupkes and Egor Lakomkin and Ehab AlBadawy and Elina Lobanova and Emily Dinan and Eric Michael Smith and Filip Radenovic and Francisco Guzmán and Frank Zhang and Gabriel Synnaeve and Gabrielle Lee and Georgia Lewis Anderson and Govind Thattai and Graeme Nail and Gregoire Mialon and Guan Pang and Guillem Cucurell and Hailey Nguyen and Hannah Korevaar and Hu Xu and Hugo Touvron and Iliyan Zarov and Imanol Arrieta Ibarra and Isabel Kloumann and Ishan Misra and Ivan Evtimov and Jack Zhang and Jade Copet and Jaewon Lee and Jan Geffert and Jana Vranes and Jason Park and Jay Mahadeokar and Jeet Shah and Jelmer van der Linde and Jennifer Billock and Jenny Hong and Jenya Lee and Jeremy Fu and Jianfeng Chi and Jianyu Huang and Jiawen Liu and Jie Wang and Jiecao Yu and Joanna Bitton and Joe Spisak and Jongsoo Park and Joseph Rocca and Joshua Johnstun and Joshua Saxe and Junteng Jia and Kalyan Vasuden Alwala and Karthik Prasad and Kartikeya Upasani and Kate Plawiak and Ke Li and Kenneth Heafield and Kevin Stone and Khalid El-Arini and Krithika Iyer and Kshitiz Malik and Kuenley Chiu and Kunal Bhalla and Kushal Lakhotia and Lauren Rantala-Yeary and Laurens van der Maaten and Lawrence Chen and Liang Tan and Liz Jenkins and Louis Martin and Lovish Madaan and Lubo Malo and Lukas Blecher and Lukas Landzaat and Luke de Oliveira and Madeline Muzzi and Mahesh Pasupuleti and Mannat Singh and Manohar Paluri and Marcin Kardas and Maria Tsimpoukelli and Mathew Oldham and Mathieu Rita and Maya Pavlova and Melanie Kambadur and Mike Lewis and Min Si and Mitesh Kumar Singh and Mona Hassan and Naman Goyal and Narjes Torabi and Nikolay Bashlykov and Nikolay Bogoychev and Niladri Chatterji and Ning Zhang and Olivier Duchenne and Onur Çelebi and Patrick Alrassy and Pengchuan Zhang and Pengwei Li and Petar Vasic and Peter Weng and Prajjwal Bhargava and Pratik Dubal and Praveen Krishnan and Punit Singh Koura and Puxin Xu and Qing He and Qingxiao Dong and Ragavan Srinivasan and Raj Ganapathy and Ramon Calderer and Ricardo Silveira Cabral and Robert Stojnic and Roberta Raileanu and Rohan Maheswari and Rohit Girdhar and Rohit Patel and Romain Sauvestre and Ronnie Polidoro and Roshan Sumbaly and Ross Taylor and Ruan Silva and Rui Hou and Rui Wang and Saghar Hosseini and Sahana Chennabasappa and Sanjay Singh and Sean Bell and Seohyun Sonia Kim and Sergey Edunov and Shaoliang Nie and Sharan Narang and Sharath Raparthy and Sheng Shen and Shengye Wan and Shruti Bhosale and Shun Zhang and Simon Vandenhende and Soumya Batra and Spencer Whitman and Sten Sootla and Stephane Collot and Suchin Gururangan and Sydney Borodinsky and Tamar Herman and Tara Fowler and Tarek Sheasha and Thomas Georgiou and Thomas Scialom and Tobias Speckbacher and Todor Mihaylov and Tong Xiao and Ujjwal Karn and Vedanuj Goswami and Vibhor Gupta and Vignesh Ramanathan and Viktor Kerkez and Vincent Gonguet and Virginie Do and Vish Vogeti and Vítor Albiero and Vladan Petrovic and Weiwei Chu and Wenhan Xiong and Wenyin Fu and Whitney Meers and Xavier Martinet and Xiaodong Wang and Xiaofang Wang and Xiaoqing Ellen Tan and Xide Xia and Xinfeng Xie and Xuchao Jia and Xuewei Wang and Yaelle Goldschlag and Yashesh Gaur and Yasmine Babaei and Yi Wen and Yiwen Song and Yuchen Zhang and Yue Li and Yuning Mao and Zacharie Delpierre Coudert and Zheng Yan and Zhengxing Chen and Zoe Papakipos},
      year={2024},
      eprint={2407.21783},
      archivePrefix={arXiv},
      primaryClass={cs.AI},
      url={https://arxiv.org/abs/2407.21783},
}

@misc{lin2024awq,
      title={AWQ: Activation-aware Weight Quantization for LLM Compression and Acceleration},
      author={Ji Lin and Jiaming Tang and Haotian Tang and Shang Yang and Wei-Ming Chen and Wei-Chen Wang and Guangxuan Xiao and Xingyu Dang and Chuang Gan and Song Han},
      year={2024},
      eprint={2306.00978},
      archivePrefix={arXiv},
      primaryClass={cs.CL},
      url={https://arxiv.org/abs/2306.00978},
}

@misc{merity2016wikitext,
      title={Pointer Sentinel Mixture Models},
      author={Stephen Merity and Caiming Xiong and James Bradbury and Richard Socher},
      year={2016},
      eprint={1609.07843},
      archivePrefix={arXiv},
      primaryClass={cs.CL},
      url={https://arxiv.org/abs/1609.07843},
}

@misc{if4,
      title={Adaptive Block-Scaled Data Types},
      author={Jack Cook and Hyemin S. Lee and Kathryn Le and Junxian Guo and Giovanni Traverso and Anantha P. Chandrakasan and Song Han},
      year={2026},
      eprint={2603.28765},
      archivePrefix={arXiv},
      primaryClass={cs.CL},
      url={https://arxiv.org/abs/2603.28765},
}

@misc{qwen2025qwen25,
      title={Qwen2.5 Technical Report},
      author={Qwen and : and An Yang and Baosong Yang and Beichen Zhang and Binyuan Hui and Bo Zheng and Bowen Yu and Chengyuan Li and Dayiheng Liu and Fei Huang and Haoran Wei and Huan Lin and Jian Yang and Jianhong Tu and Jianwei Zhang and Jianxin Yang and Jiaxi Yang and Jingren Zhou and Junyang Lin and Kai Dang and Keming Lu and Keqin Bao and Kexin Yang and Le Yu and Mei Li and Mingfeng Xue and Pei Zhang and Qin Zhu and Rui Men and Runji Lin and Tianhao Li and Tianyi Tang and Tingyu Xia and Xingzhang Ren and Xuancheng Ren and Yang Fan and Yang Su and Yichang Zhang and Yu Wan and Yuqiong Liu and Zeyu Cui and Zhenru Zhang and Zihan Qiu},
      year={2025},
      eprint={2412.15115},
      archivePrefix={arXiv},
      primaryClass={cs.CL},
      url={https://arxiv.org/abs/2412.15115},
}

@misc{raffel2020c4,
      title={Exploring the Limits of Transfer Learning with a Unified Text-to-Text Transformer},
      author={Colin Raffel and Noam Shazeer and Adam Roberts and Katherine Lee and Sharan Narang and Michael Matena and Yanqi Zhou and Wei Li and Peter J. Liu},
      year={2023},
      eprint={1910.10683},
      archivePrefix={arXiv},
      primaryClass={cs.LG},
      url={https://arxiv.org/abs/1910.10683},
}

@misc{touvron2023llama,
      title={LLaMA: Open and Efficient Foundation Language Models},
      author={Hugo Touvron and Thibaut Lavril and Gautier Izacard and Xavier Martinet and Marie-Anne Lachaux and Timothée Lacroix and Baptiste Rozière and Naman Goyal and Eric Hambro and Faisal Azhar and Aurelien Rodriguez and Armand Joulin and Edouard Grave and Guillaume Lample},
      year={2023},
      eprint={2302.13971},
      archivePrefix={arXiv},
      primaryClass={cs.CL},
      url={https://arxiv.org/abs/2302.13971},
}

@misc{vanbaalen2024gptvq,
      title={GPTVQ: The Blessing of Dimensionality for LLM Quantization},
      author={Mart van Baalen and Andrey Kuzmin and Ivan Koryakovskiy and Markus Nagel and Peter Couperus and Cedric Bastoul and Eric Mahurin and Tijmen Blankevoort and Paul Whatmough},
      year={2025},
      eprint={2402.15319},
      archivePrefix={arXiv},
      primaryClass={cs.LG},
      url={https://arxiv.org/abs/2402.15319},
}

@misc{shao2023omniquant,
      title={OmniQuant: Omnidirectionally Calibrated Quantization for Large Language Models},
      author={Wenqi Shao and Mengzhao Chen and Zhaoyang Zhang and Peng Xu and Lirui Zhao and Zhiqian Li and Kaipeng Zhang and Peng Gao and Yu Qiao and Ping Luo},
      year={2024},
      eprint={2308.13137},
      archivePrefix={arXiv},
      primaryClass={cs.LG},
      url={https://arxiv.org/abs/2308.13137},
}

@misc{kim2023squeezellm,
      title={SqueezeLLM: Dense-and-Sparse Quantization},
      author={Sehoon Kim and Coleman Hooper and Amir Gholami and Zhen Dong and Xiuyu Li and Sheng Shen and Michael W. Mahoney and Kurt Keutzer},
      year={2024},
      eprint={2306.07629},
      archivePrefix={arXiv},
      primaryClass={cs.CL},
      url={https://arxiv.org/abs/2306.07629},
}

@misc{tseng2024quipsharp,
      title={QuIP\#: Even Better LLM Quantization with Hadamard Incoherence and Lattice Codebooks},
      author={Albert Tseng and Jerry Chee and Qingyao Sun and Volodymyr Kuleshov and Christopher De Sa},
      year={2024},
      eprint={2402.04396},
      archivePrefix={arXiv},
      primaryClass={cs.LG},
      url={https://arxiv.org/abs/2402.04396},
}

@misc{eval-harness,
  author       = {Gao, Leo and Tow, Jonathan and Abbasi, Baber and Biderman, Stella and Black, Sid and DiPofi, Anthony and Foster, Charles and Golding, Laurence and Hsu, Jeffrey and Le Noac'h, Alain and Li, Haonan and McDonell, Kyle and Muennighoff, Niklas and Ociepa, Chris and Phang, Jason and Reynolds, Laria and Schoelkopf, Hailey and Skowron, Aviya and Sutawika, Lintang and Tang, Eric and Thite, Anish and Wang, Ben and Wang, Kevin and Zou, Andy},
  title        = {The Language Model Evaluation Harness},
  month        = 07,
  year         = 2024,
  publisher    = {Zenodo},
  version      = {v0.4.3},
  doi          = {10.5281/zenodo.12608602},
  url          = {https://zenodo.org/records/12608602}
}

@misc{qwen2026qwen35,
  title  = {Qwen3.5},
  author = {{Qwen Team}},
  year   = {2026},
  howpublished = {\url{https://huggingface.co/Qwen/Qwen3.5-9B}},
}
\bibliographystyle{icml2026}

\newpage
\appendix
\onecolumn

\section{$g{=}16$ Results on Qwen2.5}
\label{app:qwen25}

Table~\ref{tab:qwen25_wiki} presents $g{=}16$ WikiText-2 results on the Qwen2.5 family. AAAC achieves the best result on all four models.

\begin{table}[H]
\caption{$g{=}16$ W4A16 perplexity on Qwen2.5 models, WikiText-2 ($\downarrow$). Best quantized result in \textbf{bold}.}
\label{tab:qwen25_wiki}
\centering
\small
\begin{tabular}{l c c c c c}
\toprule
Model & BF16 & RTN & IF4 & AWQ & AAAC \\
\midrule
Qwen2.5-0.5B   & 13.03 & 14.33 & 14.11 & 14.17 & \textbf{13.66} \\
Qwen2.5-7B     & 6.80 &  7.03 &  6.97 &  7.00 & \textbf{6.93} \\
Qwen2.5-14B    & 5.25 &  5.59 &  5.54 &  5.54 & \textbf{5.45} \\
Qwen2.5-32B    & 4.97 &  5.17 &  5.12 &  5.15 & \textbf{5.09} \\
\midrule
\emph{Average}  & \emph{7.51} & \emph{8.03} & \emph{7.94} & \emph{7.97} & \emph{\textbf{7.78}} \\
\emph{Recovery}  & \emph{---} & \emph{---} & \emph{18.4\%} & \emph{12.6\%} & \emph{\textbf{47.8\%}} \\
\bottomrule
\end{tabular}
\end{table}

Table~\ref{tab:qwen25_c4} presents the corresponding C4 results. The same pattern holds across all models.

\begin{table}[H]
\caption{$g{=}16$ W4A16 perplexity on Qwen2.5 models, C4 ($\downarrow$). Best quantized result in \textbf{bold}.}
\label{tab:qwen25_c4}
\centering
\small
\begin{tabular}{l c c c c c}
\toprule
Model & BF16 & RTN & IF4 & AWQ & AAAC \\
\midrule
Qwen2.5-0.5B   & 20.02 & 21.82 & 21.57 & 21.54 & \textbf{20.95} \\
Qwen2.5-7B     & 11.77 & 12.06 & 12.02 & 11.99 & \textbf{11.93} \\
Qwen2.5-14B    & 10.25 & 10.52 & 10.49 & 10.45 & \textbf{10.39} \\
Qwen2.5-32B    & 10.09 & 10.23 & 10.21 & 10.21 & \textbf{10.18} \\
\midrule
\emph{Average}  & \emph{13.03} & \emph{13.66} & \emph{13.57} & \emph{13.55} & \emph{\textbf{13.36}} \\
\emph{Recovery}  & \emph{---} & \emph{---} & \emph{13.6\%} & \emph{17.6\%} & \emph{\textbf{47.2\%}} \\
\bottomrule
\end{tabular}
\end{table}

\section{$g{=}16$ C4 Results}
\label{app:c4_g16}

Table~\ref{tab:g16_c4} presents C4 perplexity for the same models as Table~\ref{tab:g16_wiki}. AAAC achieves the lowest perplexity on all models, with 50.9\% average gap recovery.

\begin{table}[H]
\caption{$g{=}16$ W4A16 perplexity on C4 ($\downarrow$). AAAC uses $S{=}g{=}16$ (zero overhead). Best quantized result in \textbf{bold}.}
\label{tab:g16_c4}
\centering
\small
\begin{tabular}{l c c c c c}
\toprule
Model & BF16 & RTN & IF4 & AWQ & AAAC \\
\midrule
Llama-3.2-1B   & 13.79 & 15.60 & 15.30 & 15.15 & \textbf{14.62} \\
Llama-3.2-3B   & 11.20 & 12.02 & 11.86 & 11.81 & \textbf{11.58} \\
Llama-3.1-8B   &  9.50 & 10.21 & 10.07 & 10.07 & \textbf{9.84} \\
\midrule
Qwen3.5-2B     & 17.58 & 18.59 & 18.40 & 18.29 & \textbf{18.13} \\
Qwen3.5-4B     & 13.99 & 14.51 & 14.37 & 14.38 & \textbf{14.28} \\
Qwen3.5-9B     & 12.19 & 12.57 & 12.53 & 12.44 & \textbf{12.38} \\
Qwen3.5-27B    & 10.47 & 10.64 & 10.61 & 10.59 & \textbf{10.55} \\
\midrule
\emph{Average}  & \emph{12.67} & \emph{13.45} & \emph{13.31} & \emph{13.25} & \emph{\textbf{13.05}} \\
\emph{Recovery}  & \emph{---} & \emph{---} & \emph{18.5\%} & \emph{26.0\%} & \emph{\textbf{50.9\%}} \\
\bottomrule
\end{tabular}
\end{table}

\section{Downstream Task Results}
\label{app:downstream}

Table~\ref{tab:downstream} reports per-task accuracy for Qwen3.5-9B and Qwen3.5-27B at $g{=}16$. All methods use the NVFP4 memory layout. Evaluations use the default few-shot settings from lm-evaluation-harness~\citep{eval-harness}.

\begin{table}[H]
\caption{$g{=}16$ W4A16 downstream task accuracy (\%$\uparrow$) on Qwen3.5-9B and Qwen3.5-27B. All methods use the NVFP4 memory layout. Best quantized average in \textbf{bold}.}
\label{tab:downstream}
\centering
\small
\setlength{\tabcolsep}{4.5pt}
\begin{tabular}{ll ccccc}
\toprule
Model & Task & BF16 & RTN & IF4 & AWQ & AAAC \\
\midrule
\multirow{8}{*}{Qwen3.5-9B}
 & HellaSwag & 78.1 & 77.3 & 77.3 & 77.3 & 77.6 \\
 & ARC-C     & 56.0 & 55.3 & 55.5 & 53.8 & 54.9 \\
 & PIQA      & 79.9 & 79.2 & 80.0 & 80.0 & 80.0 \\
 & BBH       & 62.3 & 60.6 & 60.1 & 61.4 & 61.7 \\
 & MuSR      & 43.4 & 42.1 & 40.5 & 44.2 & 44.2 \\
 & GPQA      & 42.2 & 39.7 & 41.9 & 41.4 & 40.0 \\
 & MATH      & 47.9 & 47.4 & 49.0 & 48.6 & 48.9 \\
 & MMLU      & 78.7 & 77.3 & 77.3 & 77.3 & 78.1 \\
\midrule
\multirow{8}{*}{Qwen3.5-27B}
 & HellaSwag & 83.4 & 83.0 & 83.2 & 83.2 & 83.2 \\
 & ARC-C     & 61.7 & 61.9 & 61.3 & 61.9 & 61.9 \\
 & PIQA      & 82.1 & 82.3 & 82.2 & 81.8 & 82.5 \\
 & BBH       & 73.1 & 72.5 & 72.2 & 72.8 & 72.3 \\
 & MuSR      & 52.5 & 53.7 & 51.3 & 52.0 & 51.6 \\
 & GPQA      & 50.1 & 49.3 & 48.2 & 49.3 & 48.5 \\
 & MATH      & 57.6 & 55.7 & 57.4 & 56.1 & 57.4 \\
 & MMLU      & 84.5 & 84.2 & 84.5 & 84.2 & 84.7 \\
\midrule
\multicolumn{2}{l}{\emph{Average}} & \emph{64.59} & \emph{63.84} & \emph{63.87} & \emph{64.08} & \emph{\textbf{64.21}} \\
\multicolumn{2}{l}{\emph{Recovery}} & \emph{---} & \emph{---} & \emph{4.0\%} & \emph{32.0\%} & \emph{\textbf{49.3\%}} \\
\bottomrule
\end{tabular}
\end{table}

\section{W4A8 Results}
\label{app:w4a8}

Although AAAC uses activations to weight importance during calibration, it is not sensitive to activation precision at inference time. The same $g{=}128$ quantized models from Table~\ref{tab:int4_awq} are evaluated with per-tensor FP8 activation quantization. Table~\ref{tab:w4a8} confirms that AAAC's improvements carry over to the W4A8 setting, with AAAC at $S{=}16$ recovering 49.1\% of the quantization gap.

\begin{table}[H]
\caption{$g{=}128$ W4A8 WikiText-2 perplexity ($\downarrow$) vs.\ AWQ. Same quantized models as Table~\ref{tab:int4_awq} with FP8 activations. \textbf{Bold}: best result per model.}
\label{tab:w4a8}
\centering
\small
\begin{tabular}{l c c c c c}
\toprule
 & & & & \multicolumn{2}{c}{AAAC} \\
\cmidrule(l){5-6}
Model & FP16 & RTN & AWQ & $S{=}128$ & $S{=}16$ \\
\midrule
Llama-3.2-3B  &  7.85 &  8.54 &  8.26 & \textbf{8.20} & \textbf{8.15} \\
Llama-3.1-8B  &  6.27 &  6.87 &  6.65 &  6.72 & \textbf{6.60} \\
Qwen2.5-0.5B  & 13.22 & 15.74 & 15.05 & \textbf{14.64} & \textbf{14.44} \\
Qwen2.5-7B    &  6.89 &  7.27 &  7.11 &  7.17 & \textbf{7.11} \\
Qwen2.5-14B   &  5.33 &  5.82 &  5.70 & \textbf{5.69} & \textbf{5.64} \\
\midrule
\emph{Average} & \emph{7.91} & \emph{8.85} & \emph{8.55} & \emph{8.48} & \emph{\textbf{8.39}} \\
\emph{Recovery} & \emph{---} & \emph{---} & \emph{31.4\%} & \emph{38.9\%} & \emph{\textbf{49.1\%}} \\
\bottomrule
\end{tabular}
\end{table}

\end{document}